\documentclass[11pt]{article}

\usepackage[preprint]{acl}

\usepackage{times}
\usepackage{latexsym}

\usepackage[T1]{fontenc}
\usepackage[utf8]{inputenc}

\usepackage{microtype}

\usepackage{inconsolata}

\usepackage{graphicx}
\usepackage{amsmath}
\usepackage{amssymb}
\usepackage{booktabs}
\usepackage{pifont}

\usepackage{caption}
\usepackage{afterpage}
\usepackage{multirow}
\usepackage{makecell}
\usepackage{bm}
\usepackage{algorithm}
\usepackage{algpseudocode}
\usepackage[most]{tcolorbox}
\definecolor{cardVisual}{HTML}{1E40AF}
\definecolor{cardExec}{HTML}{0F766E}
\definecolor{cardVerif}{HTML}{B45309}

\title{SciDiagramEdit: Learning to Edit Scientific Diagrams \\ from Paper Revisions}

\author{
  Yasheng Sun\textsuperscript{1},
  Zezi Zeng\textsuperscript{2},
  Yifan Yang\textsuperscript{2,*},
  Chong Luo\textsuperscript{2},
  Wenyi Wang\textsuperscript{1},
  Ziwei Liu\textsuperscript{3},
  J\"urgen Schmidhuber\textsuperscript{1} \\
  \textsuperscript{1}King Abdullah University of Science and Technology \quad
  \textsuperscript{2}Microsoft Research \quad
  \textsuperscript{3}Nanyang Technological University \\
  \texttt{yasheng.sun@kaust.edu.sa} \quad \texttt{yifanyang@microsoft.com} \\
  \textsuperscript{*}Corresponding author
}

\begin{document}
\makeatletter
\twocolumn[%
  \@maketitle
  \vspace{0.4em}%
  \centerline{\includegraphics[width=\textwidth]{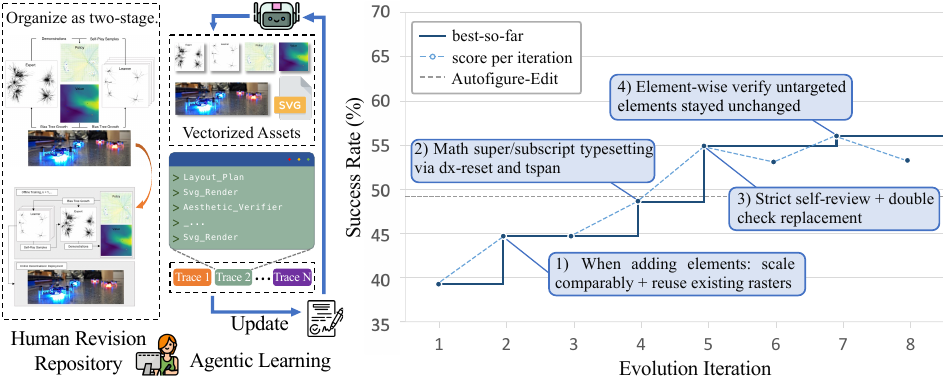}}%
  \vspace{0.4em}%
  \captionof{figure}{Overview of \textbf{SciDiagramEdit}.
    \textbf{Left:} the \emph{Human Revision Repository}, naturally
    occurring before/after figure pairs mined from arXiv paper
    revisions, where each pair encodes an edit the original authors
    performed during manuscript refinement.
    \textbf{Middle:} \emph{Agentic Learning via Skill Evolution} ---
    an SVG-editing agent edits the source figure, and execution
    traces are distilled into an evolving \emph{skill specification}
    that drives future editing behaviour.
    \textbf{Right:} success rate over evolution iterations on a
    held-out validation split.  The best-so-far skill snapshot
    rises monotonically; callouts illustrate representative
    learned skill rules.}%
  \label{fig:teaser}%
  \vspace{1em}%
]
\makeatother

\begin{abstract}
Editing the figures in a research paper is a routine and
time-consuming part of everyday research practice: authors relabel
components, rearrange panels, and restyle visuals as they revise
their manuscripts.
Automating this editing workflow under a natural-language
instruction, however, is challenging, because a scientific figure
is a dense infographic in which heterogeneous visual elements such
as schematics, plots, photos, captions, and arrows are composed
under a tight visual grammar to advance a specific argument.
To address this, we present \textbf{SciDiagramEdit}, a benchmark
and skill-evolution framework that learns from natural paper
revisions and operates on the figure's editable vector source,
where users can inspect and co-edit individual primitives alongside
the agent.
Our benchmark mines before/after figure pairs from arXiv version
histories, each grounded in the authors' own revision intent.
To accommodate the diversity of editing instructions, we adopt
\emph{agentic learning via skill evolution}: an agentic proposer
continually refines the agent's skill specification from execution
traces over multiple epochs.
Experiments show that the agent learns meaningful editing skills
from paper revisions and substantially improves edit quality over
strong single-pass baselines.
\end{abstract}

\section{Introduction}
\label{sec:intro}

Scientific figures often determine whether a paper's contribution
is grasped at a glance, and researchers refine them through many
cycles of revision before camera-ready submission: relabelling
components, rearranging panels, restyling visuals.  These edits
are typically driven by natural-language feedback from coauthors
and reviewers, making instruction-driven figure editing one of the
most common figure-related operations in everyday research
practice.  Existing automated tools, however, target diagram \emph{generation}
from text~\citep{zhu2026autofigure, zhu2026paperbanana,
huang2026scifig} and its
evaluation~\citep{liao2026aibench, chang2025sridbench}, offering
little support for the editing regime.

Automating instruction-driven editing of scientific figures
presents a central challenge: a scientific figure is a dense
infographic in which heterogeneous visual elements such as
schematics, plots, photos, captions, and arrows are composed under
a tight visual grammar to advance a specific argument.  Editing such a figure therefore demands localised,
compositional operations on its primitives that combine many
sub-skills, making one-shot hand-engineered scaffolds brittle
across the long tail of edit types.
Existing commercial text-to-image editors such as Nano
Banana Pro~\citep{google2025nanobanana} and
GPT-Image-2~\citep{openai2025gptimage} can re-render a complex
figure from a natural-language prompt, but they operate at the
raster level in a single pass and expose no compositional handle
for the targeted local operations a researcher actually performs
during revision, such as adding or removing a sub-panel,
relabelling a specific component, or re-routing arrows.  Natural
paper revisions, which capture genuine author-written edit intent
at scale, remain a largely untapped source of supervision for
scientific-figure editing.

To address this challenge, we present \textbf{SciDiagramEdit}, a
skill-evolution framework that learns to edit scientific diagrams
from natural paper revisions.
\emph{Our key insight is that author-drawn figure revisions encode
the implicit visual grammar of scientific communication, and that
an agent can distil this grammar by contrasting each of its edits
against the author-drawn target.}  This insight plays
out in two parts.  First, the publication record itself already
contains a long trail of natural editing pairs: between two arXiv
versions of the same paper, the same figure is often re-rendered
by the authors.  Mining these pairs yields a benchmark of
before/after figures against which a rubric-grounded judge scores
an agent's edit by directly comparing it to the author-drawn
target.  Second, because the space of editing instructions is too
diverse for any fixed scaffold to anticipate, we adopt
\emph{agentic learning via skill evolution}: by contrasting each
rollout against the author-drawn target, an agentic proposer
continually distils the drawing patterns observed in those targets
into the agent's skill specification over multiple epochs, lifting
edit accuracy without requiring a hand-engineered pipeline.  To
implement this loop, the editor operates directly on the figure's
editable vector source, which has the additional benefit of keeping
every primitive open for users to inspect and co-edit alongside the
agent.

Our contributions are threefold:
\textbf{1)}~We introduce \textsc{SciDiagramEdit}, an editing
benchmark of 364~before/after figure pairs mined from real arXiv
paper revisions across 23 subjects, annotated with 2{,}628 atomic
editing claims that capture the authors' own revision intent.
\textbf{2)}~We build a self-improving scientific figure-editing
agent that learns from paper revisions via skill evolution,
distilling execution traces and the authors' demonstrations into
a portable skill specification.
\textbf{3)}~Experiments show that our agent learns meaningful
editing skills from paper revisions and substantially improves
edit quality over strong single-pass baselines.

\section{Related Work}
\label{sec:related}

\subsection{Scientific Diagram Generation}
\label{sec:rw-generation}

A first line of work generates a scientific diagram from a textual
description.  Recent systems include
\textsc{AutoFigure}~\citep{zhu2026autofigure},
\textsc{PaperBanana}~\citep{zhu2026paperbanana}, and
\textsc{SciFig}~\citep{huang2026scifig}, which build on general SVG
generation backbones such as
\textsc{StarVector}~\cite{rodriguez2025starvector},
\textsc{LLM4SVG}~\cite{xing2025empowering}, and
\textsc{OmniSVG}~\cite{yang2025omnisvgunifiedscalablevector}.
A parallel line evaluates such systems:
\textsc{AIBench}~\cite{liao2026aibench} and
\textsc{SridBench}~\cite{chang2025sridbench} adopt rubric-based
protocols whose checklist style inspires the judge we use for the
editing setting.  All of these systems and benchmarks address
diagrams produced \emph{from scratch}; none takes an existing
diagram plus an editing instruction as input.

% Float placed here (before §2.2) so that it lands on page 3 top
% rather than being deferred deeper into the document.  Its caption
% still belongs conceptually to §3 (Dataset Curation).
\begin{table*}[!t]
\centering\small
\begin{tabular*}{\linewidth}{@{\extracolsep{\fill}}llllr@{}}
\toprule
\textbf{Benchmark} & \textbf{Task} & \textbf{Domain} & \textbf{Source} & \textbf{\#} \\
\midrule
\multicolumn{5}{@{}l}{\emph{Scientific-illustration evaluation}} \\
AIBench~\citep{liao2026aibench}              & Eval          & Sci illustration     & Papers, top venues 2025    &    300   \\
SridBench~\citep{chang2025sridbench}         & Eval          & Sci illustration     & Papers, 13 disciplines     &  1{,}120 \\
\midrule
\multicolumn{5}{@{}l}{\emph{Scientific-illustration generation}} \\
SVGEditBench~V2~\citep{nishina2025svgeditbench}  & Edit          & Generic SVG (emojis) & Emoji icons, GPT-4o prompts &  1{,}683 \\
AutoFigure~\citep{zhu2026autofigure}               & Gen           & Sci illustration     & Papers, surveys, blogs      &  3{,}300 \\
AutoFigure-Edit~\citep{lin2026autofigure}      & Gen           & Sci illustration     & Papers, AutoFigure subset   &    200   \\
PaperBananaBench~\citep{zhu2026paperbanana}        & Gen           & Methodology diag.    & Papers, NeurIPS 2025        &    292   \\
\textbf{SciDiagramEdit (ours)} & \textbf{Edit} & \textbf{Sci illustration} & \textbf{Papers, arXiv revisions} & \textbf{364} \\
\bottomrule
\end{tabular*}
\caption{Benchmarks adjacent to our setting, all spanning 2024--2026.  \textbf{Task}: \emph{Gen} $=$ text$\to$figure generation; \emph{Eval} $=$ evaluation of generated figures; \emph{Edit} $=$ instruction-grounded edit of an existing figure.  Only \textsc{SciDiagramEdit} combines instruction-driven editing with scientific diagrams sourced from real author-written paper revisions.}
\label{tab:benchmarks}
\end{table*}

\subsection{Scientific Diagram Editing}
\label{sec:rw-editing}

The work closest to ours is
\textsc{AutoFigure-Edit}~\citep{lin2026autofigure}, a one-shot
pipeline that converts a raw illustration into an editable SVG but
does not itself \emph{learn how to edit} given an instruction.
\textsc{Chat2SVG}~\citep{wu2025chat2svg} performs iterative
instruction-driven editing on generic SVGs.  In the raster domain,
\textsc{InstructPix2Pix}~\citep{brooks2023instructpix2pix} and
commercial editors such as Nano
Banana Pro~\citep{google2025nanobanana} and
GPT-Image-2~\citep{openai2025gptimage}, together with
native-unified VLMs such as
\textsc{SenseNova-U1}~\citep{diao2026sensenova}, re-render the
image in a single pass and expose no addressable vector primitives
for targeted local edits.  Among editing benchmarks,
\textsc{SVGEditBench V2}~\cite{nishina2025svgeditbench} and
\textsc{SVGenius}~\cite{chen2025svgenius} evaluate instruction-based
SVG editing on generic content, with neither targeting scientific
diagrams.

\subsection{Self-Improving LLM Agents}
\label{sec:rw-self-improving}

A line of work improves LLM-agent behaviour through
natural-language feedback over execution traces, varying in the
\emph{unit} of evolution.
\textsc{TextGrad}~\cite{yuksekgonul2024textgradautomaticdifferentiationtext} operates at the
level of textual gradients through a compound AI system;
\textsc{Voyager}~\citep{wang2023voyager},
\textsc{GEPA}~\citep{agrawal2025gepa},
\textsc{Trace2Skill}~\citep{ni2026trace2skill}, and
\textsc{EvoSkill}~\citep{alzubi2026evoskill} evolve prompts and
skill artifacts; and
\textsc{Meta-Harness}~\citep{lee2026meta} evolves the
surrounding coding harness.
We adopt the same patch-and-merge philosophy and specialise it to
scientific-diagram editing, where the unit of evolution is a skill
specification consumed by an SVG-editing agent operating on
editable vector primitives.

\section{Dataset Curation}
\label{sec:dataset}

\begin{figure*}[!t]
  \centering
  \includegraphics[width=\linewidth]{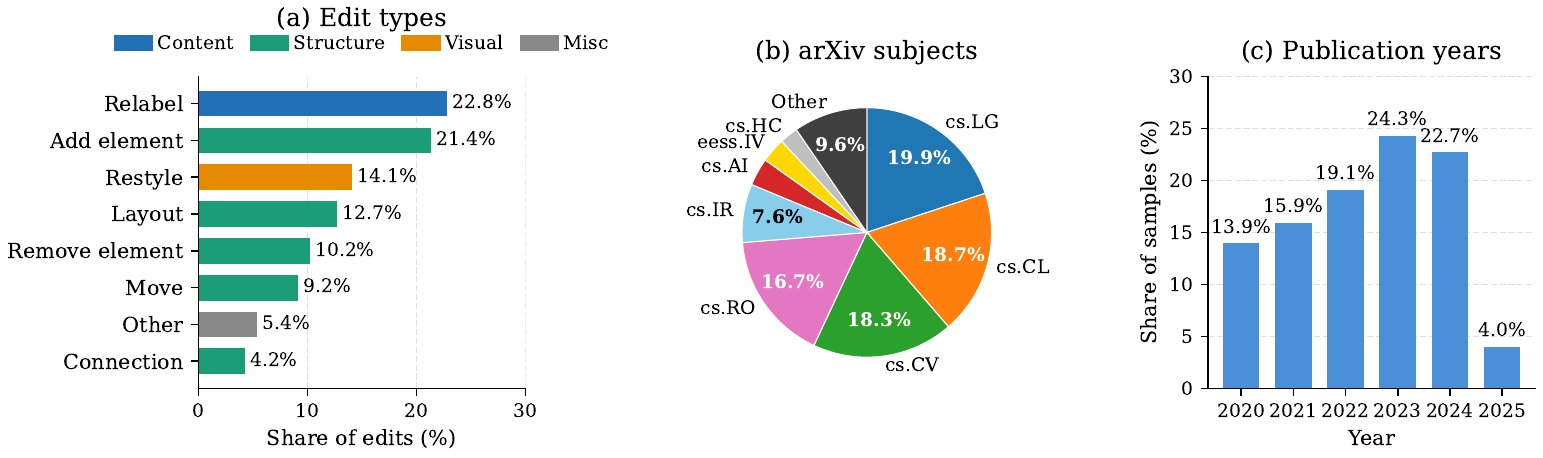}
  \caption{Dataset statistics.
    (a)~Edit-type distribution over 2{,}628 atomic claims across the
    364~samples, grouped into Content / Structure / Visual / Misc
    semantic categories.
    %  renaming and element addition together account for 44\% of all edits.
    (b)~Source papers span 23~arXiv primary categories;
    the four machine-learning categories (\textsc{cs.LG},
    \textsc{cs.CL}, \textsc{cs.CV}, \textsc{cs.RO}) jointly cover
    73.6\% of samples, with the remainder distributed across a long
    tail.
    (c)~Publication years skew toward recent revisions.}
  \label{fig:dataset-stats}
\end{figure*}

We construct \textbf{SciDiagramEdit}, a benchmark of 364~natural
before/after figure-edit pairs mined from arXiv paper revisions.
Each pair places the same figure across two arXiv versions
($v_{\mathrm{old}}$ and $v_{\mathrm{new}}$) of a paper, providing
a naturally occurring edit grounded in real research practice.

\paragraph{Relation to Prior Benchmarks.}
Table~\ref{tab:benchmarks} situates \textsc{SciDiagramEdit} relative
to adjacent benchmarks.  Among scientific-illustration benchmarks
(FigureBench, PaperBananaBench, AIBench, SridBench), all target
\emph{generation} or \emph{evaluation} of figures produced from
scratch and lack paired before/after data.  Among instruction-based
SVG editing benchmarks (SVGEditBench~V2, AutoFigure-Edit), only
SVGEditBench~V2 provides paired natural-language-instructed edits,
but it operates on generic emoji SVGs rather than scientific
diagrams.  Our benchmark is the only one combining (i)~the editing
regime, (ii)~scientific-diagram domain, (iii)~paired before/after
figures, and (iv)~naturally occurring author-written instructions,
with paper \emph{revisions} as the data source.

\paragraph{Distribution and Diversity.}
Figure~\ref{fig:dataset-stats}\,(a) breaks down edits into four
semantic groups: \textbf{Content} (relabeling text),
\textbf{Structure} (adding, removing, moving, re-laying-out, or
re-routing elements), \textbf{Visual} (restyling colours, fonts,
strokes), and \textbf{Misc}.  Content-level renaming
(\textsc{Relabel}, 22.8\%) and structural additions
(\textsc{Add element}, 21.4\%) jointly account for 44\% of edits,
while connection rewiring is the rarest operation (4.2\%).
Figure~\ref{fig:dataset-stats}\,(b) summarises the source-paper
distribution: papers span 23~arXiv primary subjects, with the
four largest categories (\textsc{cs.LG}, \textsc{cs.CL},
\textsc{cs.CV}, \textsc{cs.RO}) accounting for 73.6\% of samples
and a long tail of 15~further categories adding disciplinary
breadth.  Sample years (Figure~\ref{fig:dataset-stats}\,c) skew
toward recent revisions, with 47\% drawn from 2023--2024.

\paragraph{Annotation.}
Scientific figures combine schematic elements that are naturally
vectorisable (arrows, frames, text labels, connectors) with raster
regions (photos, complex plots, dense sub-schematics) that resist
lossless vectorisation.  To enable instruction-driven editing at
the primitive level, we decompose each figure into these two
layers via the pipeline of
\textsc{AutoFigure-Edit}~\citep{lin2026autofigure}.  (i)~A
region segmenter identifies the raster panels and returns a set
of bounding boxes covering them.  (ii)~An LLM writes an SVG
template that vectorises the surrounding schematic elements as
native SVG primitives, leaving a placeholder slot at each
bounding box.  (iii)~The original raster crop of each panel is
inserted back into its placeholder as an embedded
\texttt{<image>} node.  Formally, we represent each
processed figure as a pair
\begin{equation}
\label{eq:figure}
F \;=\; (V,\, A), \quad A \;=\; \{a_k\}_{k=1}^{m},
\end{equation}
where $V$ is the SVG source (vectorised primitives plus
\texttt{<image>} reference nodes) and $A$ is the set of $m$
embedded raster panels indexed by placeholder.

Beyond the figure pair itself, each sample is annotated at
curation time with a natural-language editing instruction and a
per-sample checklist
\begin{equation}
\label{eq:checklist-set}
Q \;=\; \big\{(q_j,\, \mathcal{C}_j,\, a^{\star}_j)\big\}_{j=1}^{n},
\end{equation}
where $q_j$ is a natural-language question grounded in the
instruction (e.g., ``Was the panel-(C) caption added below the
rightmost block?''), $\mathcal{C}_j$ is the admissible-answer set
(binary $\{\textsc{yes},\,\textsc{no}\}$ or a multi-choice list),
and $a^{\star}_j\!\in\!\mathcal{C}_j$ is the ground-truth answer.
Across the 364 samples we collect 2{,}628 such questions in total
(Figure~\ref{fig:dataset-stats}\,a).  The questions are designed
to cover both surface-level instruction-following (whether each
named change appears in the output) and the figure's overall
logical coherence (whether the edit preserves the diagram's
reasoning structure).

\section{Methodology}
\label{sec:method}

\begin{figure*}[!t]
  \centering
  \includegraphics[width=\linewidth]{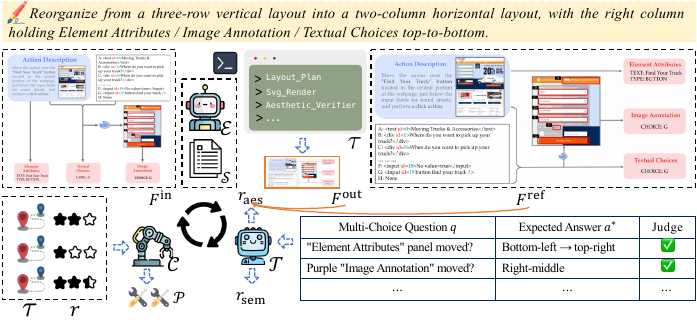}
  \caption{Overview of the \textbf{SciDiagramEdit} training loop on
    a representative sample.  Given an input figure $F^{\mathrm{in}}$
    and a natural-language editing instruction, the
    \textbf{Editor}~$\mathcal{E}$ uses the current skill
    specification $\mathcal{S}$ to produce an edited figure
    $F^{\mathrm{out}}$ and an execution trace $\tau$.  The
    \textbf{Judge}~$\mathcal{J}$ scores the output along two axes:
    an aesthetic preference $r_{\mathrm{aes}}$ from a pairwise
    comparison against the author's revision $F^{\mathrm{ref}}$,
    and a semantic faithfulness $r_{\mathrm{sem}}$ from a per-sample
    multiple-choice checklist.  The \textbf{Coach}~$\mathcal{C}$
    consumes the trace, the scores, and $F^{\mathrm{ref}}$ as a
    demonstration, emitting a patch that updates $\mathcal{S}$.
    The loop iterates over training minibatches from
    $\mathcal{D}_{\mathrm{train}}$ with validation gating on
    $\mathcal{D}_{\mathrm{val}}$.}
  \label{fig:method}
\end{figure*}

\subsection{Problem Formulation}
\label{sec:problem}

Given an editable scientific figure
$F^{\mathrm{in}}\!=\!(V^{\mathrm{in}}, A^{\mathrm{in}})$ and a
natural-language editing instruction $I$, the goal is to produce a
figure that satisfies $I$ while preserving figure content unrelated
to the instruction.

\paragraph{Architecture Overview.}
Our architecture (Figure~\ref{fig:method}) centres on the
\textbf{Editor}~$\mathcal{E}$, an agentic system that uses
code-level tools to perform the various edits on the SVG figure
under a skill specification $\mathcal{S}$.  To progressively improve
the Editor's ability across instructions, we further introduce a
\textbf{Judge}~$\mathcal{J}$ that evaluates the Editor's current
performance, and a \textbf{Coach}~$\mathcal{C}$ that
reads past traces and Judge verdicts to upgrade the Editor by
emitting patches to $\mathcal{S}$.
Iterating this loop refines $\mathcal{S}$ across training steps.

\subsection{Scientific Diagram Editor $\mathcal{E}$}
\label{sec:evolve}

The Editor is an agentic system, realised as a code-writing
subprocess with file-system and Python tool access.  Given
$(F^{\mathrm{in}}, I, \mathcal{S})$, it issues a sequence of tool
calls $(a_t, o_t)_{t=1}^{T}$ that manipulate the SVG, yielding an
edited figure $F^{\mathrm{out}}$ and an execution trace
$\tau=(a_t, o_t)_{t=1}^{T}$.  The Editor loads the top-level
\texttt{SKILL.md} verbatim into its system prompt and reads any
referenced \texttt{workflows/} or \texttt{tools/} file on demand
via its file-system tool, so context cost is independent of
the skill directory's size.

\paragraph{Skill Specification.}
Formally, $\mathcal{S}$ is a finite map from relative paths to file
contents,
\begin{equation}
\label{eq:skill}
\mathcal{S} \;=\; \big\{(p_i,\, c_i)\big\}_{i=1}^{|\mathcal{S}|}, \quad p_i\!\in\!\mathcal{R},\; c_i\!\in\!\Sigma^{\ast},
\end{equation}
where $\mathcal{R}$ is the set of admissible relative paths (a
top-level \texttt{SKILL.md} plus optional \texttt{workflows/} and
\texttt{tools/} subdirectories), and $\Sigma^{\ast}$ denotes
arbitrary unicode content.  Following recent work on agentic
skills~\citep{alzubi2026evoskill, ni2026trace2skill}, we treat
$\mathcal{S}$ as a learnable artefact and evolve it from execution
traces rather than authoring it by hand.

\paragraph{Tool Primitives.}
To save the Editor from reinventing common utilities and to
shorten its exploration, we expose several optional CLI primitives
that it can call on demand: \texttt{render-svg} provides
headless-Chromium SVG rendering with full SVG/CSS/font support,
\texttt{layout-lint} runs deterministic geometric checks on the
rendered output, and \texttt{gen-icon} synthesises small raster
icons of real-world objects (e.g.\ a sensor, a server rack) that
SVG primitives cannot naturally draw.  When and how to invoke
these primitives is encoded in $\mathcal{S}$.

\subsection{Aesthetic-Aware Judge $\mathcal{J}$}
\label{sec:judge}

The Judge is a vision-language model that assigns a scalar score
$r\!\in\![0,1]$ to each edit, combining a
\emph{semantic-faithfulness} term that quantifies how well the
edit realises the changes specified in the instruction with an
\emph{aesthetic-preference} term that gates the score on whether
the output matches the target's visual polish.

\paragraph{Semantic Faithfulness.}
We assess instruction-following, factual, and logical fidelity at
sub-instruction granularity through the per-sample checklist $Q$
of Eq.~\ref{eq:checklist-set} rather than free-form scoring:
constraining the VLM to a small admissible answer set raises
answer-level precision and yields an interpretable per-question
breakdown, in line with the question-answering protocol used by
recent academic-illustration
evaluators~\citep{liao2026aibench}.  Given an edited figure
$F^{\mathrm{out}}$, the VLM is queried with
$(F^{\mathrm{out}}, F^{\mathrm{in}}, F^{\mathrm{ref}}, I, q_j)$
and returns a predicted answer
$\hat{a}_j\!\in\!\mathcal{C}_j$ for each question, yielding the
verification accuracy
\begin{equation}
\label{eq:checklist-pass}
r_{\mathrm{sem}} \;=\; \frac{1}{n}\sum_{j=1}^{n}\mathbb{1}\!\left[\hat{a}_j = a^{\star}_j\right] \,\in\,[0,1].
\end{equation}

\paragraph{Aesthetic Preference.}
Pointwise VLM scoring of a single figure's aesthetic quality is
known to be noisy, whereas pairwise comparison between two
figures yields a more reliable preference
signal~\citep{wang2025pref}.  We therefore evaluate
aesthetics by asking the VLM to compare the Editor's output
$F^{\mathrm{out}}$ directly against the author's revision
$F^{\mathrm{ref}}$ in randomised order to control position bias,
yielding a binary preference
\begin{equation}
\label{eq:aesthetic-gate}
r_{\mathrm{aes}} \;=\; \mathbb{1}\!\left[F^{\mathrm{out}} \succeq F^{\mathrm{ref}}\right],
\end{equation}
where $\succeq$ denotes ``judged at least as polished as'' under
the comparison.

\paragraph{Composite Score.}
An edit is only useful in practice if it is both semantically
faithful and visually presentable.  We therefore gate verification
accuracy on aesthetic preference, rather than averaging the two
components:
\begin{equation}
\label{eq:reward}
r \;=\; r_{\mathrm{aes}} \cdot r_{\mathrm{sem}},
\end{equation}
so an edit must reach the target's aesthetic threshold before any
verification credit accrues to it.

\subsection{Demonstration-Aware Coach $\mathcal{C}$}
\label{sec:proposer}

% Source Table 2 here (early in §4.4 which lives on p6) so that
% the [!t] float can claim the top of p6.
\begin{table*}[!t]
\centering\small
\begin{tabular*}{\linewidth}{@{\extracolsep{\fill}}lccccc@{}}
\toprule
\multicolumn{1}{c}{\multirow{2}{*}{\textbf{Method}}}
& \multicolumn{2}{c}{\textbf{Semantic}}
& \multicolumn{3}{c}{\textbf{Aesthetic}} \\
\cmidrule(lr){2-3}\cmidrule(lr){4-6}
  & Success Rate\,$\uparrow$
  & Win Rate\,$\uparrow$
  & IAA\,$\uparrow$
  & ISTA\,$\uparrow$
  & Win Rate\,$\uparrow$ \\
\midrule
\textsc{GPT-Image-1.5}~\citep{openai2025gptimage}      & 0.516         & 0.347         & \textbf{50.77}   & \textbf{46.60}   & 0.349             \\
\textsc{GPT-Image-2}~\citep{openai2025gptimage}        & \underline{0.882}& \textbf{0.778}& \underline{49.60}& \underline{45.78}& \textbf{0.637}    \\
\midrule
\textsc{AutoFigure-Edit} (\textsc{GPT-5.3})~\citep{lin2026autofigure} & 0.828         & 0.528         & 45.30         & 41.24         & 0.389             \\
\textsc{AutoFigure-Edit} (\textsc{GPT-5.4})~\citep{lin2026autofigure} & 0.823         & 0.520         & 45.08         & 41.35         & 0.317             \\
\textsc{AutoFigure-Edit} (\textsc{GPT-5.5})~\citep{lin2026autofigure} & 0.844         & 0.573         & 44.87         & 41.37         & 0.400             \\
\midrule
\textbf{Ours}                      & \textbf{0.932}& \underline{0.756}& 47.93            & 45.58            & \underline{0.515} \\
\midrule
\textit{Target (GT)}              & ---           & ---           & \textit{48.25}& \textit{46.49}& ---            \\
\bottomrule
\end{tabular*}
\caption{Quantitative comparison on the test set.  Both win
rates are blind-pairwise versus the author's revised figure.
\textbf{Semantic}: checklist success rate and
instruction-following win rate.  \textbf{Aesthetic}:
UniPercept's Image Aesthetic Assessment (IAA) and Image
Structure and Texture Assessment (ISTA) on a 0--100
scale~\citep{liao2026aibench}, and aesthetic win rate.
Higher is better; \emph{Target (GT)} reports UniPercept on
the author's revision.}
\label{tab:main-results}
\end{table*}

The Coach progressively improves the Editor's ability by editing
the skill specification $\mathcal{S}$ in light of past Editor
trajectories.  We realise it as a coding-agent
subprocess with file-system and shell access, dropped into a
workspace that exposes the current skill directory $\mathcal{S}$,
the per-sample trajectory bundles, the author's revised figure
$F^{\mathrm{ref}}$, and a feedback-history file $H$ summarising
prior accept/reject decisions so previously rejected ideas are not
re-litigated.

\paragraph{Patch Operator.}
Let $\mathcal{B}\!\subset\!\mathcal{D}_{\mathrm{train}}$ be a
minibatch of $M$ training samples, with $\bm{\tau},\,\bm{r}$
the per-sample traces and judge scores produced by the Editor on
$\mathcal{B}$.  The Coach outputs a patch of at most $L$ edits,
\begin{equation}
\label{eq:patch}
\mathcal{P} \;=\; \mathcal{C}\bigl(\mathcal{S},\, \mathcal{B},\, \bm{\tau},\, \bm{r},\, H\bigr),
\end{equation}
where each edit
$\pi = (\textsc{action}, p, \textsc{op}, c)$ specifies an action
(\textsc{create\_file} or \textsc{edit\_file}), a target path $p$,
an operator \textsc{op} drawn from standard file-modification
primitives (\textsc{create}, \textsc{rewrite}, \textsc{append},
\textsc{insert\_after}, \textsc{replace}, \textsc{delete}), and
new content $c$.  The schema lets a single edit introduce a new
\texttt{workflows/} file rather than append to
\texttt{SKILL.md}, keeping the always-loaded core concise as
expertise accumulates.

\paragraph{Training Loop.}
\label{sec:trainloop}
The Coach is invoked iteratively on minibatches drawn from
$\mathcal{D}_{\mathrm{train}}$ and each candidate is scored on a
held-out validation split $\mathcal{D}_{\mathrm{val}}$.  We maintain a
top-$K$ frontier of the best-scoring skills seen so far and propose
new candidates from frontier entries in round-robin order; once training
terminates, the highest-scoring entry of the frontier is returned:
\begin{equation}
\label{eq:objective}
\mathcal{S}^{\star} \;=\; \arg\max_{\mathcal{S}}\; \bar{r}\!\left(\mathcal{S};\, \mathcal{D}_{\mathrm{val}}\right),
\end{equation}
where the argmax ranges over candidates seen during training and
$\bar{r}(\mathcal{S}; \mathcal{D}) :=
\mathbb{E}_{(F^{\mathrm{in}}, I, F^{\mathrm{ref}}) \sim \mathcal{D}}
\bigl[\mathcal{J}(\mathcal{E}(F^{\mathrm{in}}, I; \mathcal{S}),
F^{\mathrm{in}}, F^{\mathrm{ref}}, I)\bigr]$ is the expected
per-sample score.\footnote{Full pseudocode appears as
Algorithm~\ref{alg:reflact} in the appendix.}

\paragraph{Beyond Trace and Score.}
The trace $\tau$ and score $r$ tell the Coach what the Editor
attempted and how well it fared, but neither shows what the edit
should look like.  The author's revised figure $F^{\mathrm{ref}}$
supplies this missing \emph{demonstration}: comparing
$F^{\mathrm{ref}}$ against the Editor's output $F^{\mathrm{out}}$
lets the Coach distil the systematic gaps into generalisable skill
rules, capturing the compositional and stylistic conventions the
author followed rather than sample-specific fixes.

\section{Experiments}
\label{sec:experiments}

\begin{table*}[!t]
\centering\small
\begin{tabular*}{\linewidth}{@{\extracolsep{\fill}}lcccccc@{}}
\toprule
\multicolumn{1}{c}{\multirow{2}{*}{\textbf{Model}}}
  & \multicolumn{2}{c}{\textit{w/o skill}}
  & \multicolumn{2}{c}{\textit{w/ evolved skill}}
  & \multicolumn{2}{c}{\textit{w/ transferred skill}} \\
\cmidrule(lr){2-3}\cmidrule(lr){4-5}\cmidrule(lr){6-7}
  & Semantic\,$\uparrow$ & Aesthetic\,$\uparrow$
  & Semantic\,$\uparrow$ & Aesthetic\,$\uparrow$
  & Semantic\,$\uparrow$ & Aesthetic\,$\uparrow$ \\
\midrule
\textsc{GPT-5.1} & 0.224        & 0.168        & 0.252\,\textsuperscript{\scriptsize $+0.028$} & 0.205\,\textsuperscript{\scriptsize $+0.037$} & 0.257\,\textsuperscript{\scriptsize $+0.033$} & 0.199\,\textsuperscript{\scriptsize $+0.031$} \\
\textsc{GPT-5.3} & 0.659        & 0.285        & 0.713\,\textsuperscript{\scriptsize $+0.054$} & 0.358\,\textsuperscript{\scriptsize $+0.073$} & 0.706\,\textsuperscript{\scriptsize $+0.047$} & 0.366\,\textsuperscript{\scriptsize $+0.081$} \\
\textsc{GPT-5.4} & 0.700        & 0.360        & 0.724\,\textsuperscript{\scriptsize $+0.024$} & 0.436\,\textsuperscript{\scriptsize $+0.076$} & 0.718\,\textsuperscript{\scriptsize $+0.018$} & 0.425\,\textsuperscript{\scriptsize $+0.065$} \\
\textsc{GPT-5.5} & 0.745        & 0.466        & 0.756\,\textsuperscript{\scriptsize $+0.011$} & 0.515\,\textsuperscript{\scriptsize $+0.049$} & ---          & ---          \\
\bottomrule
\end{tabular*}
\caption{Effect of skill across backbones.  Each column
group reports semantic and aesthetic win rates under one
skill setting: \emph{w/o skill}, \emph{w/ evolved skill},
and \emph{w/ transferred skill} (the skill evolved with
\textsc{GPT-5.5} reused here).  Superscripts on the
\emph{w/ skill} columns show the change ($\Delta$) relative
to the \emph{w/o skill} baseline in the same row.}
\label{tab:transfer}
\end{table*}

\begin{figure*}[!t]
  \centering
  \includegraphics[width=\linewidth]{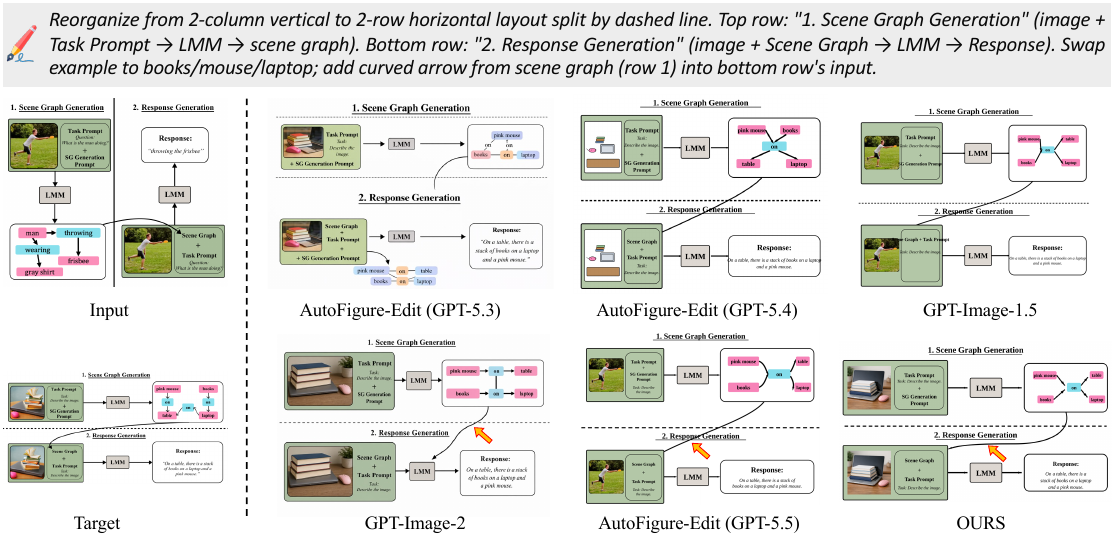}
  \caption{Qualitative comparison on a representative editing
    instance.  The panels present the input figure, the author's
    target revision, and the outputs of baseline editors
    (\textsc{AutoFigure-Edit} with \textsc{GPT-5.3}, \textsc{GPT-5.4},
    \textsc{GPT-5.5}; \textsc{GPT-Image-1.5}; \textsc{GPT-Image-2})
    alongside our method (\textbf{Ours}).}
  \label{fig:quali}
\end{figure*}

\subsection{Experiment Setup}
\label{sec:exp-setup}

\paragraph{Evaluation Protocol.}
We measure the quality of the agent's edits along two axes.
\textbf{Semantic Faithfulness} is scored by the per-sample
checklist success rate~$r_{\mathrm{sem}}$ and a blind-pairwise
instruction-following win rate.
\textbf{Aesthetic Quality} is scored by \textsc{UniPercept}'s
Image Aesthetic Assessment (IAA) and Image Structure and Texture
Assessment (ISTA) on a 0--100 scale~\citep{liao2026aibench}, and
a blind-pairwise aesthetic win rate~$r_{\mathrm{aes}}$.  Both
pairwise comparisons use the author's revised figure as the
reference.
All scores are computed on the test split of the
\textsc{SciDiagramEdit} corpus (\S\ref{sec:dataset}), split
$2{:}1{:}3$ into
$\mathcal{D}_{\mathrm{train}},\, \mathcal{D}_{\mathrm{val}},\, \mathcal{D}_{\mathrm{test}}$;
skill evolution runs for two epochs on
$\mathcal{D}_{\mathrm{train}}$, with candidates accepted only
when they do not regress the running validation score on
$\mathcal{D}_{\mathrm{val}}$.
% Full hyperparameters are in Appendix~\ref{sec:appendix}.

\paragraph{Comparison Approaches.}
We compare against two families of approaches drawn from
Figure~\ref{fig:quali}.
(i)~\textbf{\textsc{AutoFigure-Edit}}-style \textbf{Editors}~\citep{lin2026autofigure}: the LLM is shown the input
SVG source and the editing instruction and emits the full edited
SVG in a single pass.  We evaluate three frontier LLM backbones
(\textsc{GPT-5.3}, \textsc{GPT-5.4}, \textsc{GPT-5.5}).
(ii)~\textbf{Raster Regenerators}: \textsc{GPT-Image-1.5}
and \textsc{GPT-Image-2}~\citep{openai2025gptimage}, which accept
the figure and instruction as a raster pair and re-render an
edited raster.  We include \textsc{GPT-Image-2}, the strongest
closed-source image editor available, as our demanding
raster-regime reference.

\begin{figure*}[!t]
  \centering
  \includegraphics[width=\linewidth]{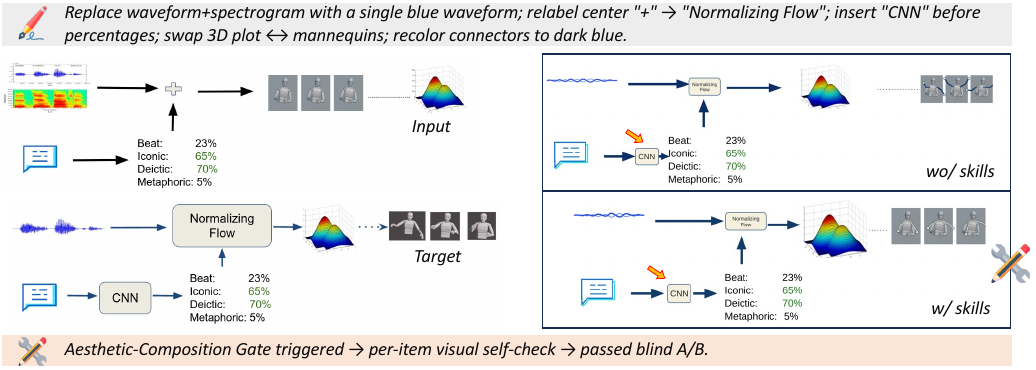}
  \vspace{-11pt}
  \caption{Qualitative skill ablation on a representative
    editing instance.  Left: the input figure (top) and the
    author's revised target (bottom); right: our method's
    outputs without the evolved skill (top) and with it
    (bottom).}
  \vspace{-7pt}
  \label{fig:skill-ablation}
\end{figure*}

\subsection{Quantitative Results}
\label{sec:exp-quantitative}

\paragraph{Comparison with Baselines.}
The raster regenerators set a high bar on the aesthetic
axis: on UniPercept's IAA and ISTA, both
\textsc{GPT-Image-1.5} and \textsc{GPT-Image-2} score
strongly, even surpassing the author's revised figure
(\emph{Target}), consistent with their position as
state-of-the-art closed-source image editors.  On the
semantic axis, however, \textbf{Ours} attains the highest
score on the semantic checklist, surpassing both raster
regenerators and all single-pass \textsc{AutoFigure-Edit}
backbones, and is essentially tied with \textsc{GPT-Image-2}
on the instruction-following win rate.  On the aesthetic axis,
\textbf{Ours} is the strongest vector-aware method and
nearly matches \emph{Target} on both IAA and ISTA, though
it trails the raster regenerators on the blind pairwise
aesthetic gate.

\paragraph{Effect of Skill Across Backbones.}
Table~\ref{tab:transfer} tests whether the procedural
knowledge captured by the evolved skill transfers across
backbones.  We hold the skill evolved on \textsc{GPT-5.5}
fixed and apply it to each backbone in turn.  The
transferred skill yields gains on both axes, with the
aesthetic axis benefiting more.  It is also comparable to
each backbone's self-evolved skill (\emph{w/ evolved skill});
we conjecture that \textsc{GPT-5.5}'s greater capacity lets
it explore a wider space of skill rules during evolution,
producing rules that generalise well to weaker backbones.

\subsection{Qualitative Results}
\label{sec:exp-qualitative}

\paragraph{Comparison Against Baselines.}
Figure~\ref{fig:quali} shows our qualitative comparison on
a scene-graph figure that the author rearranged from a
2-column vertical layout into a 2-row horizontal layout with
a dashed separator and a curved arrow leading from the row-1
scene graph back into the row-2 input.
\textsc{GPT-Image-2} trades logical fidelity for visual
polish: the rendering is clean, but the curved arrow lands
on the wrong target panel.  The
\textsc{AutoFigure-Edit} variants make the opposite trade,
getting the semantic content of the instruction right but at
the cost of aesthetic quality, with the new connector
crossing the surrounding label text.  \textbf{Ours} navigates
this trade-off: the curved arrow lands on the correct target
panel, and the connector does not cross the surrounding
label text.\footnote{See the appendix for additional visual
examples.}

\paragraph{Effect of the Evolved Skill.}
Figure~\ref{fig:skill-ablation} compares our method's
outputs without (\emph{wo/ skills}, top right) and with
(\emph{w/ skills}, bottom right) the evolved skill on a
single editing instance.  Without the skill, the inserted
CNN box and the percentage labels are crammed against each
other and the panel reads as visually cramped.  With the
skill in place, the \emph{Aesthetic-Composition Gate} rule
(annotated in orange) triggers a per-item visual self-check
that rebalances the inter-element spacing into a more
breathable composition.

\subsection{User Study}
\label{sec:user-study}

Beyond the automated scores, we run a human pairwise study
against the two strongest baselines, \textsc{GPT-Image-2}
and \textsc{AutoFigure-Edit} (\textsc{GPT-5.5}).  We sample 30
instances uniformly from $\mathcal{D}_{\mathrm{test}}$ and
recruit five volunteer participants who each rate
\textsc{Ours} against each baseline along two axes,
\emph{aesthetic} and \emph{instruction following}, in
separate rounds.  Each trial shows the input figure
$F^{\mathrm{in}}$, the natural-language instruction, and two
anonymised outputs in randomised left/right order; the
author's revision $F^{\mathrm{ref}}$ is withheld.
Table~\ref{tab:userstudy} reports the win rate of
\textsc{Ours} against each baseline along each axis.

\begin{table}[t]
\centering\small
\begin{tabular*}{\linewidth}{@{\extracolsep{\fill}}lcc@{}}
\toprule
\textbf{\textsc{Ours} vs.}
  & \textbf{Aesthetic}
  & \textbf{Semantic} \\
\midrule
\textsc{GPT-Image-2}                            & 0.54 & 0.59 \\
\textsc{AutoFigure-Edit}\,(\textsc{GPT-5.5})    & 0.63 & 0.68 \\
\bottomrule
\end{tabular*}
\vspace{-6pt}
\caption{We report the win rate of \textsc{Ours} against each
baseline on 30 instances sampled from
$\mathcal{D}_{\mathrm{test}}$, with five participants under a
blinded protocol.}
\vspace{-10pt}
\label{tab:userstudy}
\end{table}

\section{Conclusion}
\label{sec:conclusion}

We presented \textsc{SciDiagramEdit}, a skill-evolution
framework for editing scientific diagrams that operates on the
figure's editable vector source and acquires a library of
editing rules from execution traces and author-drawn targets,
so its outputs are SVG the user can continue editing, not
a flattened raster.  Alongside the framework, we
release a benchmark mined from arXiv paper revisions, where
each before/after pair is grounded in the authors' own
revision intent.  On this benchmark, the agent matches the strongest
closed-source raster editor on semantic faithfulness while
keeping the figure's structure editable, and its learned
skill further transfers across backbones.

\section*{Limitations}

Our benchmark focuses on editing instructions whose intent is
stated explicitly: each edit clause is spelled out in the
natural-language prompt.  We have not yet explored edits that
require multi-step reasoning over the figure's underlying
argument, nor the harder regime where the agent must infer
authorial intent from the surrounding paper context rather
than from the prompt itself.  In the same spirit, the
skill-evolution loop in this work is still externally
orchestrated; a more product-grade pipeline would self-refine
autonomously, weighing research taste alongside the explicit
metric signal and progressively reducing human supervision in
the loop.  Finally, the experiments we report operate at a
relatively modest scale, with a few hundred training pairs
and a few dozen evolution steps.  Scaling either axis
substantially, by mining many more revision pairs or running
the loop for many more iterations, may surface emergent agent
behaviours and more abstract skill rules than the current
compute budget affords.

\section*{Ethical Considerations}

\paragraph{Data sources and licensing.}
The benchmark is mined from publicly available arXiv
preprints.  The underlying papers and figures retain their
original licences and we use them under fair use for
non-commercial research.  We will release the curated
instruction--revision pairs and our atomic-claim annotations
under \textsc{CC BY-NC 4.0} for research use only; the source
figures themselves are not redistributed and must be fetched
from arXiv under each paper's original licence.

\paragraph{Content and privacy.}
All figures in the benchmark are technical scientific
diagrams such as plots, scene graphs, and network or pipeline
diagrams.  During curation we manually inspected every pair
and did not encounter personally identifying information
beyond the author bylines already present in arXiv metadata,
which we do not include in the model inputs.  We also did not
encounter offensive content.

\paragraph{Reproducibility.}
Because the framework relies on closed-weight
\textsc{GPT-5.x} backbones, training-time access is gated by
API cost.  We release the benchmark and the skill
specification produced by the evolution loop so that future
work can reuse the distilled rules without re-running
training.

\paragraph{Potential for misuse.}
Like any figure-editing tool, our system could in principle
be used to misrepresent experimental visuals.  Its intended
use is to support legitimate authoring during revision, where
the author retains editorial authority and every
model-suggested modification is left as inspectable SVG
primitives rather than baked into pixels.

\paragraph{AI assistants.}
The \textsc{GPT-5.x} and \textsc{Claude} models are integral
components of our method.  Separately, we used an AI
assistant in two human-supervised roles.  First, during
dataset curation, as a co-author of the natural-language
editing instruction, with every released annotation
human-confirmed.  Second, for prose polishing of the paper
itself, namely paraphrasing and grammar refinement of
author-written passages, without it suggesting new technical
content.

% \section*{Acknowledgments}
%
% \emph{[Acknowledgments placeholder --- suppressed in review mode.]}

% Bibliography entries for the entire Anthology, followed by custom entries
%\bibliography{custom,anthology-overleaf-1,anthology-overleaf-2}

% Custom bibliography entries only
\bibliography{custom}

\appendix

\section*{Appendix}

\section{Additional Technical Details}

\subsection{Annotation Interface}
\label{sec:appendix-annotation}

\begin{figure*}[!htbp]
  \centering
  \includegraphics[width=\linewidth]{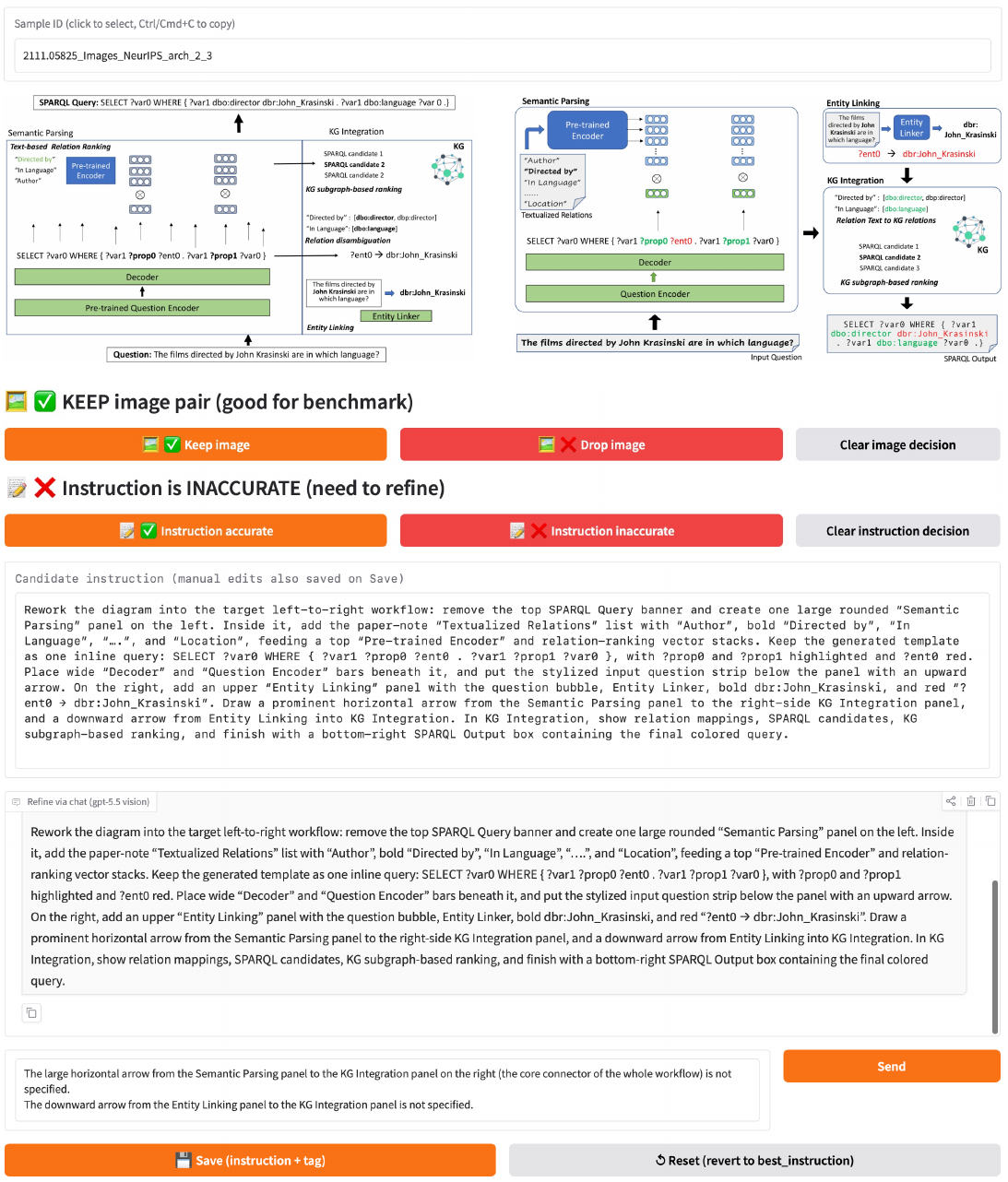}
  \caption{Gradio interface used to annotate the
    natural-language editing instruction for each
    \textsc{SciDiagramEdit} pair.  The before/after figure
    pair is shown side-by-side, followed by binary verdicts
    on whether to keep the pair and whether the candidate
    instruction is accurate, an editable instruction field,
    and a chat panel for iteratively refining the instruction
    with a vision LLM.}
  \label{fig:annotation-ui}
\end{figure*}

Raw paired-figure mining from arXiv revisions is noisy: two
versions of the same figure often differ only in a re-rendered
raster or a font swap with no semantic intent, and a small
fraction are wholesale replacements that share no structure
with the original.  We keep only the meaningful middle band,
in which the author has added a panel, renamed a label,
re-routed an arrow, or re-laid out a block of elements while
retaining enough of the original scaffold for the change to
read as an edit rather than a redraw.  Even within this band
the raw visual diff seldom makes the author's intent
explicit on its own: the same surface change can encode
different revision intents, and disambiguating which is meant
requires reading the surrounding paper context.  A useful
editing instruction therefore has to be authored, not just
extracted, which is what Figure~\ref{fig:annotation-ui} is
designed to support.

The interface in Figure~\ref{fig:annotation-ui} covers the
instruction-annotation stage of curation.  The before/after
figure pair is rendered at the top of the panel.  The curator
first applies two filtering decisions: a binary keep/drop
verdict on whether the pair is suitable for the benchmark,
and a binary accurate/inaccurate verdict on whether the
candidate instruction faithfully describes the visual edit.
The candidate instruction itself is shown in an editable text
field that the curator can revise directly.  Below it, a chat
panel backed by a vision LLM, \textsc{GPT-5.5}, lets
the curator iteratively refine the instruction by describing
what is missing or wrong; the assistant returns a revised
draft that the curator can accept, edit further, or discard.
The loop typically converges in two to three rounds.  The
\emph{Save} button commits the final instruction together
with the curator's tags; the \emph{Reset} button reverts to
the last saved version.  Two parallel panels share this layout but are not shown for
brevity: one filters SVG vectorisations whose rendering
drifts noticeably from the source raster, since the
\textsc{AutoFigure-Edit} vectoriser of \S\ref{sec:dataset}
occasionally produces SVG that diverges visually from the
original figure and we drop those samples; the other
authors the per-sample checklist of
Eq.~\ref{eq:checklist-set}.  All editing instructions and
checklist questions are written in English; the source
figures are drawn from English-language arXiv preprints.

\subsection{Skill Evolution}
\label{sec:appendix-settings}

\paragraph{Training Loop.}
Algorithm~\ref{alg:reflact} formalises the skill-evolution
loop sketched in \S\ref{sec:trainloop}.  The loop maintains
a top-$K$ frontier $\mathcal{F}$ that stores the
highest-scoring skills seen so far, in the spirit of the
Pareto-frontier search of \citet{alzubi2026evoskill}.  Each
iteration picks a parent from $\mathcal{F}$ in round-robin
order, runs the Editor on a fresh training minibatch with
that skill, lets the Judge score the resulting outputs, and
asks the Coach to write a patch from the traces, scores, and
the running feedback history $H$.  The patched skill is
re-evaluated on $\mathcal{D}_{\mathrm{val}}$ and admitted to
$\mathcal{F}$ whenever the frontier still has room or its
score strictly exceeds the frontier's worst entry, in which
case that worst entry is evicted; otherwise the candidate is
rejected.  Either way the accept/reject verdict is appended
to $H$ so the Coach does not re-propose a rejected idea.

\begin{algorithm}[h]
\caption{Skill-Evolution Training Loop}
\label{alg:reflact}
\begin{algorithmic}[1]
\Require Splits $\mathcal{D}_{\mathrm{train}}, \mathcal{D}_{\mathrm{val}}$; initial skill $\mathcal{S}_0$; Editor $\mathcal{E}$, Judge $\mathcal{J}$, Coach $\mathcal{C}$; total steps $T$; minibatch size $M$; edit budget $L$; frontier size $K$
\Ensure Best-so-far skill $\mathcal{S}^{\star}$
\State $s_0 \gets \bar{r}(\mathcal{S}_0;\, \mathcal{D}_{\mathrm{val}})$
\State Initialise frontier $\mathcal{F} \gets \{(\mathcal{S}_0,\, s_0)\}$,\, $H \gets \emptyset$
\For{$t = 1, \ldots, T$}
  \State $(\mathcal{S},\, s) \gets$ next entry of $\mathcal{F}$ in round-robin order \Comment{round-robin parent}
  \State Sample minibatch $\mathcal{B} = \{(F^{\mathrm{in}}_i, I_i, F^{\mathrm{ref}}_i)\}_{i=1}^{M} \sim \mathcal{D}_{\mathrm{train}}$
  \For{$i = 1, \ldots, M$}
    \State $(F^{\mathrm{out}}_i, \tau_i) \gets \mathcal{E}(F^{\mathrm{in}}_i, I_i;\, \mathcal{S})$
    \State $r_i \gets \mathcal{J}(F^{\mathrm{out}}_i, F^{\mathrm{in}}_i, F^{\mathrm{ref}}_i, I_i)$
  \EndFor
  \State $\mathcal{P} \gets \mathcal{C}(\mathcal{S},\, \mathcal{B},\, \bm{\tau},\, \bm{r},\, H)$ \Comment{$|\mathcal{P}|\le L$}
  \State $\tilde{\mathcal{S}} \gets$ apply $\mathcal{P}$ to $\mathcal{S}$
  \State $\tilde s \gets \bar{r}(\tilde{\mathcal{S}};\, \mathcal{D}_{\mathrm{val}})$
  \If{$|\mathcal{F}| < K$} \Comment{room in the frontier}
    \State $\mathcal{F} \gets \mathcal{F} \cup \{(\tilde{\mathcal{S}}, \tilde s)\}$
    \State $H \gets H \cup \{(\mathcal{P}, \textsc{accept}, \tilde s)\}$
  \ElsIf{$\tilde s > \min_{(S, s') \in \mathcal{F}} s'$} \Comment{beats current worst}
    \State $\mathcal{F} \gets \bigl(\mathcal{F} \setminus \{\arg\min_{(S, s') \in \mathcal{F}} s'\}\bigr) \cup \{(\tilde{\mathcal{S}}, \tilde s)\}$
    \State $H \gets H \cup \{(\mathcal{P}, \textsc{accept}, \tilde s)\}$
  \Else
    \State $H \gets H \cup \{(\mathcal{P}, \textsc{reject}, \tilde s)\}$
  \EndIf
\EndFor
\State $(\mathcal{S}^{\star},\, s^{\star}) \gets \arg\max_{(S, s') \in \mathcal{F}} s'$
\State \Return $\mathcal{S}^{\star}$
\end{algorithmic}
\end{algorithm}

\paragraph{Settings.}
Table~\ref{tab:settings} lists the concrete configuration used
to instantiate this loop in our main experiments, and a few
choices there are worth motivating.  The Editor uses a
\textsc{GPT-5.x} backbone while the Coach and Judge use
\textsc{Claude Opus 4.7}: the cross-family pairing keeps the
Editor from gaming a same-family Judge's idiosyncratic
preferences, Claude Opus's code competence suits the Coach's
role of authoring skill-rule patches, and the capacity gap
between a stronger Coach and a weaker Editor lets procedural
knowledge distilled by the Coach lift students of varying
capacity, including \textsc{GPT-5.1}, \textsc{5.3}, and
\textsc{5.4}.  The search itself is multi-frontier:
evolution maintains a top-$K{=}3$ frontier of candidate skills
rather than a single best-so-far, since a frontier of one
plateaued quickly in early experiments while multi-frontier
search consistently escaped local optima.  The reward
governing admission to that frontier is the multiplicative
composite $r = r_{\mathrm{aes}} \cdot r_{\mathrm{sem}}$ from
Eq.~\ref{eq:reward} rather than an average; because
vector-aware editing tends to under-perform on aesthetics,
multiplying rather than averaging raises the bar there,
gating out semantically faithful edits whose aesthetics still
fall short and steering the loop to focus on that axis.
Skill updates are applied as patches over the prior skill
rather than as full rewrites, which empirically avoids
forgetting rules the Coach had previously introduced for
unrelated failure modes.

\begin{table}[!htbp]
\centering\small
\begin{tabular*}{\linewidth}{@{\extracolsep{\fill}}ll@{}}
\toprule
\textbf{Component} & \textbf{Value} \\
\midrule
\multicolumn{2}{l}{\emph{Models}} \\
Editor $\mathcal{E}$ (student)  & \textsc{GPT-5.5} \\
Editor backbones (transfer)     & \textsc{GPT-5.1/5.3/5.4/5.5} \\
Coach $\mathcal{C}$ (analyst)   & \textsc{Claude Opus 4.7} (1M\,ctx) \\
Judge $\mathcal{J}$ (vision)    & \textsc{Claude Opus 4.7} \\
\midrule
\multicolumn{2}{l}{\emph{Evolution loop}} \\
Frontier size                   & 3 (multi-frontier) \\
Frontier strategy               & round-robin parent \\
Epochs                          & 2 \\
Batch size                      & 8 \\
Max analyst rounds per step     & 3 \\
Edit budget per skill update    & 8 patches \\
Skill update mode               & patch (vs.\ rewrite) \\
Initial skill scaffold          & basic tool usage and guide \\
\midrule
\multicolumn{2}{l}{\emph{Reward}} \\
Composite $r$ (Eq.~\ref{eq:reward}) & $r_{\mathrm{aes}} \cdot r_{\mathrm{sem}}$ \\
Semantic $r_{\mathrm{sem}}$         & checklist accuracy $\in [0,1]$ \\
Aesthetic $r_{\mathrm{aes}}$        & binary, pairwise vs.\ target \\
\midrule
\multicolumn{2}{l}{\emph{Inference}} \\
Coach rewrite max tokens        & 64{,}000 \\
Coach parallelism               & 4 workers \\
Editor rollout parallelism      & 8 workers \\
Editor call timeout             & 600\,s \\
\bottomrule
\end{tabular*}
\caption{Settings used for the skill-evolution loop and the
results reported in \S\ref{sec:experiments}.}
\label{tab:settings}
\end{table}

\paragraph{Compute Resources.}
All model inference is performed through hosted APIs; we do
not train any model weights and do not use any GPU compute.
The Editor calls \textsc{GPT-5.x} through the \textsc{Codex}
CLI, and the Coach and Judge call \textsc{Claude Opus 4.7}
through the \textsc{Claude Code} CLI.  Orchestration, SVG
rendering, and trajectory bookkeeping run on a single
CPU-only workstation.  Across all reported experiments in
this paper, spanning four Editor backbones together with the
skill-evolution, transfer, and ablation runs and the
development cycles that preceded them, total API spend was on
the order of US\,\$20{,}000 at standard rate-card pricing.  We use
headless \textsc{Chromium} for SVG\,$\to$\,PNG rendering, the
\textsc{Codex} CLI for the Editor, the \textsc{Claude Code}
CLI for the agentic Coach, and the official \texttt{anthropic}
SDK for the Judge.

\subsection{User Study Protocol}
\label{sec:appendix-userstudy}

The user study reported in \S\ref{sec:user-study} was
conducted with five volunteer participants from the authors'
institution; all participants are graduate students in
computer science / electrical engineering and participated
without compensation.  Before each session, participants were
informed verbally that their anonymised preferences would be
aggregated and reported in a research paper, and they
consented before proceeding.  The protocol collects only
anonymous A/B preferences over already-public scientific
figures, does not record any personally identifying
information, and was determined to be exempt from formal IRB
review by our institution on that basis.

Each participant judges every one of the 30 sampled instances
against each of the two baselines on each of the two axes,
giving $30 \times 2 \times 2 = 120$ forced-choice trials per
participant and 600 trials across the five participants.  The
trials are presented in two rounds.  In the aesthetic round,
for each instance the participant sees two trials in
randomised left/right order: one pairs \textsc{Ours} against
\textsc{GPT-Image-2} and one pairs \textsc{Ours} against
\textsc{AutoFigure-Edit} (\textsc{GPT-5.5}); for each trial
the participant answers \emph{``Which output is more visually
polished?''}.  The semantic round repeats the same
per-instance setup with the question \emph{``Which output
more faithfully realises the instruction?''}.  Each trial
displays the input figure $F^{\mathrm{in}}$, the
natural-language editing instruction, and the two anonymised
outputs; the author's revision $F^{\mathrm{ref}}$ is withheld
throughout, and no ``tie'' option is provided.
Table~\ref{tab:userstudy} reports the proportion of votes in
favour of \textsc{Ours} per baseline and per axis, computed
over the resulting 150 votes per cell.

\section{Further Analysis}

\subsection{Typical Learned Skills}
\label{sec:appendix-skills}

To give a concrete sense of what the skill-evolution loop
actually produces, we first show one evolved skill at work,
then describe the broader library that the final skill pack
$\mathcal{S}^{\star}$ contains.

\begin{figure*}[!t]
  \centering
  \includegraphics[width=\linewidth]{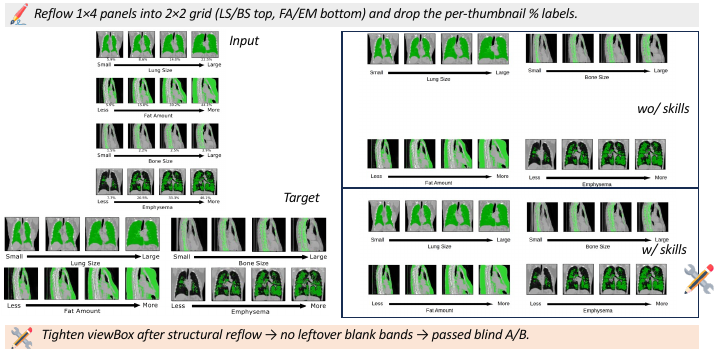}
  \caption{Effect of an evolved skill.  The instruction
    reflows a 1$\times$4 panel grid into a 2$\times$2 layout
    and drops the per-thumbnail percentage labels.  Without the skill, the Editor leaves
    leftover blank bands around the new content; with the
    skill, the learned rule ``tighten the \texttt{viewBox}
    after structural reflow'' eliminates them and the output
    passes the blind aesthetic A/B against the author's
    target.}
  \label{fig:skill-equip}
\end{figure*}

Figure~\ref{fig:skill-equip} shows one such skill in action.
The \emph{wo/ skills} output gets the panel reflow right but
the SVG \texttt{viewBox} still encloses the original
1$\times$4 footprint, so visible bands surround the new
content and the aesthetic gate fails.  With the evolved
skill, the rule ``tighten the \texttt{viewBox} after any
structural reflow'' fires and the bands disappear.

Beyond this single rule, the pack is not a single,
general-purpose prompt: the agent instead accumulates a
structured library of trigger-plus-rule entries, each
targeting a specific recurring failure mode the Coach observed
in Editor traces.  On disk the pack has the following layout:

{\small
\begin{verbatim}
SKILL/
|-- SKILL.md
`-- workflows/
    |-- safe-string-replace.md
    |-- math-notation.md
    |-- post-restructure-compactness.md
    |-- deliverable.md
    |-- workspace.md
    |-- available_cli_tools_call_via_bash.md
    `-- ...
\end{verbatim}
}

\noindent The top-level \texttt{SKILL.md} groups rules by their
trigger; longer sub-routines are factored out into individual
files under \texttt{workflows/}.  Below we highlight five
representative entries that span the abstraction range: a
top-level verification loop, a visual-fidelity rule, two
concrete edit-execution idioms covering string replacement and
math typesetting, and a structural-edit verification rule whose
entry doubles as a meta-comment on the typical shape committed
rules take.  Two patterns recur across these entries.  The
verification rules cover distinct scopes, with a whole-image
check after every edit and a separate scoped check after
structural edits, rather than collapsing into a single
``check everything'' rule.  The concrete edit idioms in turn
target SVG-specific traps, such as overlapping string
replacements, \texttt{<tspan>} ordering, and \texttt{viewBox}
tightening, that a generic ``be careful'' instruction would
never surface.

\begin{tcolorbox}[
  enhanced, breakable,
  colback=cardVerif!4!white, colframe=cardVerif,
  colbacktitle=cardVerif, coltitle=white,
  fonttitle=\bfseries\small,
  title={Render-and-Verify Loop\hfill \scriptsize\textsc{Verification}},
  boxrule=0.4pt, arc=2pt,
  left=4pt, right=4pt, top=4pt, bottom=4pt,
  before skip=4pt, after skip=4pt
]
\small
\textbf{Trigger.}\ Every editing task, before writing
\texttt{summary.txt}.\\[2pt]
\textbf{Failure.}\ A single ``looks done'' judgment from the
Editor is the most common path to a failed eval: shipped
outputs either miss an instruction clause or carry an aesthetic
regression the Editor did not notice in the SVG source.\\[2pt]
\textbf{Rule.}\ Always render \texttt{output.png} and pass it
through a strict vision-LLM reviewer that returns a JSON
\texttt{ship}/\texttt{revise} verdict alongside an enumerated
list of missing clauses and aesthetic issues; if the verdict is
\texttt{revise}, fix the listed items and re-render before
writing the summary.  Never accept a single optimistic
self-check.
\end{tcolorbox}

\begin{tcolorbox}[
  enhanced, breakable,
  colback=cardVisual!4!white, colframe=cardVisual,
  colbacktitle=cardVisual, coltitle=white,
  fonttitle=\bfseries\small,
  title={Match the Visual Register\hfill \scriptsize\textsc{Visual fidelity}},
  boxrule=0.4pt, arc=2pt,
  left=4pt, right=4pt, top=4pt, bottom=4pt,
  before skip=4pt, after skip=4pt
]
\small
\textbf{Trigger.}\ Edits that add a new element (icon, label,
panel, callout) or rewrite a notation that recurs across the
figure.\\[2pt]
\textbf{Failure.}\ Technically-correct additions still lose the
blind A/B when they do not visually belong: under-sized so they
read as decoration rather than as a diagram element, or drawn
as crude SVG primitives where the source uses polished raster
or vector icons.\\[2pt]
\textbf{Rule.}\ Aim for an output that \emph{could plausibly be
the published version} of the input.  Anchor a new element's
size to a comparable existing element's bounding box, and
reuse the source's representation (the existing
\texttt{<image>} reference, or the matching \texttt{<g>}
subtree anchored by a distinctive fill colour) when the
instruction names a known icon.  Surface correctness is
necessary but not sufficient.
\end{tcolorbox}

\begin{tcolorbox}[
  enhanced, breakable,
  colback=cardExec!4!white, colframe=cardExec,
  colbacktitle=cardExec, coltitle=white,
  fonttitle=\bfseries\small,
  title={Anchored String Replace\hfill \scriptsize\textsc{Edit execution}},
  boxrule=0.4pt, arc=2pt,
  left=4pt, right=4pt, top=4pt, bottom=4pt,
  before skip=4pt, after skip=4pt
]
\small
\textbf{Trigger.}\ Edits that swap two labels
(\texttt{A}$\leftrightarrow$\texttt{B}) or rewrite a label that
recurs across symmetric panels (e.g.,
\textsc{MAML} vs.\ \textsc{Ours}, before vs.\ after).\\[2pt]
\textbf{Failure.}\ A naive \texttt{Path.read\_text()
{\textrightarrow} .replace() {\textrightarrow} write\_text()}
idiom has two sharp edges: \texttt{.replace(A,B)} followed by
\texttt{.replace(B,A)} undoes the first edit, and a bare
\texttt{.replace(label,new)} silently rewrites the label in
\emph{both} panels when only one was named.\\[2pt]
\textbf{Rule.}\ For two-way swaps, route both sides through a
\texttt{@@hash@@} placeholder pass.  For symmetric-panel
relabels, anchor each \texttt{.replace()} with the element's
\texttt{x="\ldots"} coordinate (or any other unique
neighbouring context) so the rewrite is panel-scoped instead of
file-wide.
\end{tcolorbox}

\begin{tcolorbox}[
  enhanced, breakable,
  colback=cardVisual!4!white, colframe=cardVisual,
  colbacktitle=cardVisual, coltitle=white,
  fonttitle=\bfseries\small,
  title={Math-Notation Typesetting\hfill \scriptsize\textsc{Visual fidelity}},
  boxrule=0.4pt, arc=2pt,
  left=4pt, right=4pt, top=4pt, bottom=4pt,
  before skip=4pt, after skip=4pt
]
\small
\textbf{Trigger.}\ Sub/superscripts on the same symbol
(e.g., $\pi_t^{\mathrm{Alice}}$, $\smash{I_u^{(\mathrm{un})p}}$)
or an inline math legend listing several notations.\\[2pt]
\textbf{Failure.}\ Sequential
\texttt{<tspan baseline-shift>} calls flow the subscript
\emph{after} the superscript rather than under it; Unicode
super/subscript codepoints render at the wrong scale and break
the surrounding font register.\\[2pt]
\textbf{Rule.}\ Emit real \texttt{<tspan>} elements with
\texttt{dx="-N\,em"} to back-shift the subscript under the
superscript, and rewrite any Unicode super/subscripts into the
same tspan idiom.
\end{tcolorbox}

\begin{tcolorbox}[
  enhanced, breakable,
  colback=cardVerif!4!white, colframe=cardVerif,
  colbacktitle=cardVerif, coltitle=white,
  fonttitle=\bfseries\small,
  title={Layered Structural Review\hfill \scriptsize\textsc{Verification}},
  boxrule=0.4pt, arc=2pt,
  left=4pt, right=4pt, top=4pt, bottom=4pt,
  before skip=4pt, after skip=4pt
]
\small
\textbf{Trigger.}\ Instruction contains any of \texttt{delete},
\texttt{remove}, \texttt{swap}, \texttt{replace \ldots\ with},
\texttt{move \ldots\ into}, \texttt{restructure},
\texttt{reorganize}.\\[2pt]
\textbf{Failure.}\ The strict whole-image reviewer judges the
output plausible and returns \texttt{ship}, but the blind A/B
against $F^{\mathrm{ref}}$ still fails on four category-specific
side-effects the holistic pass misses: duplicated content where
a move was requested, blank regions left after a deletion,
shrunken replacements, and wrong-coloured replacements.\\[2pt]
\textbf{Rule.}\ Chain a \emph{scoped} second vision-LLM pass
(\texttt{workflows/post-\allowbreak{}restructure-\allowbreak{}compactness.md})
that audits only those four side-effect categories under a JSON
schema; treat each non-empty category as a separate
\texttt{revise} signal even when the first pass already
shipped.\\[4pt]
\emph{Why this is the typical shape.}\ The Coach commits the
specific reviewer-shipping pattern (whole-image reviewer +
structural-edit triggers + four named side-effects) rather than
the abstract principle ``check layout after structural edits.''
Across iterations, committed rules follow this shape: a precise
trigger paired with a scoped sub-routine, not a general
exhortation.
\end{tcolorbox}

\subsection{Demonstration-Aware Coach Ablation}
\label{sec:appendix-demo-aware}

A central design choice for the Coach is whether to expose
the author's revised figure $F^{\mathrm{ref}}$ as a worked
demonstration, or to rely on the trajectories alone.  We
ablate this choice by comparing a \emph{demonstration-aware}
Coach that sees $F^{\mathrm{ref}}$ against a
\emph{demonstration-free} Coach that does not, with
everything else about the loop held fixed.

Both variants are \textsc{Claude Code} subprocesses with
filesystem access to each sample folder; the analyst uses its
\texttt{Bash} and \texttt{Read} tools to inspect trajectories
and writes proposed patches to
\texttt{analyst\_response.json}.  They differ only in whether
the author's target files are in the workspace: the
demonstration-aware variant places \texttt{target.png} and
\texttt{target.svg} alongside
\texttt{input.png}/\texttt{output.png} so the analyst can
\texttt{Read} them as multimodal input, while the
demonstration-free variant omits the targets, leaving only
the input/output renders and trajectory transcripts.  Their
system prompts also differ:

\begin{tcolorbox}[
  enhanced, breakable,
  colback=cardVerif!4!white, colframe=cardVerif,
  colbacktitle=cardVerif, coltitle=white,
  fonttitle=\bfseries\small,
  title={Demonstration-Free Analyst},
  boxrule=0.4pt, arc=2pt,
  left=4pt, right=4pt, top=4pt, bottom=4pt,
  before skip=4pt, after skip=4pt
]
\small
``You will be given MULTIPLE failed agent trajectories from a
single minibatch and the current skill document.  Your job is
to identify the most important COMMON failure patterns across
the batch and propose a concise set of skill edits.''  The
analyst has filesystem access to \texttt{input.png} and
\texttt{output.png} for each sample via its \texttt{Read}
tool; only the author's target files
(\texttt{target.png}, \texttt{target.svg}) are withheld from
the workspace.
\end{tcolorbox}

\begin{tcolorbox}[
  enhanced, breakable,
  colback=cardExec!4!white, colframe=cardExec,
  colbacktitle=cardExec, coltitle=white,
  fonttitle=\bfseries\small,
  title={Demonstration-Aware Analyst (with $F^{\mathrm{ref}}$)},
  boxrule=0.4pt, arc=2pt,
  left=4pt, right=4pt, top=4pt, bottom=4pt,
  before skip=4pt, after skip=4pt
]
\small
Each sample folder is described to the analyst as containing
\texttt{input.png} (rendered input), \texttt{output.png}
(rendered output), \texttt{target.svg} (ground-truth SVG
authored as the edit), and \texttt{target.png} (rendered
ground truth).  The prompt then instructs:
\begin{enumerate}\itemsep0pt
\item Read both PNGs \texttt{input.png}, \texttt{output.png},
  and \texttt{target.png}: \emph{``your Read tool supports
  images; see visually what changed and what didn't.''}
\item Examine \texttt{target.svg} source: \emph{``how did
  the ground-truth author solve the same edit?''}
\item Decompose what makes \texttt{target.svg} into skill
  rules.
\end{enumerate}
\end{tcolorbox}

\begin{figure*}[!t]
  \centering
  \includegraphics[width=\linewidth]{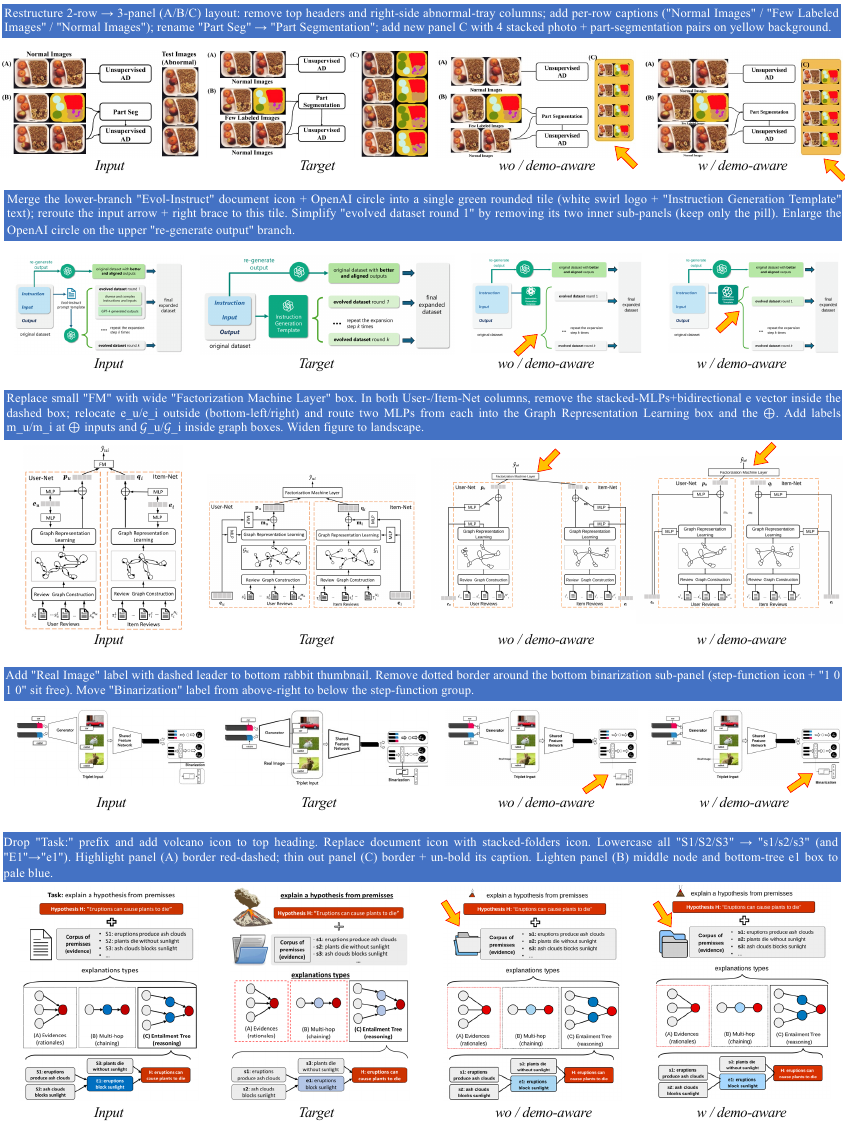}
  \caption{Qualitative ablation of the demonstration-aware
    Coach (\S\ref{sec:proposer}).  Each row presents an
    editing instruction with its input figure, the author's
    revised target, the Editor's output under a skill evolved
    \emph{without} access to the author's revision
    ($F^{\mathrm{ref}}$ withheld from the Coach;
    \emph{wo/ demo-aware}), and the Editor's output under our
    full demonstration-aware Coach (\emph{w/ demo-aware}).
    Orange arrows mark places where the demonstration-aware
    variant realises an instruction detail that the
    demonstration-free variant misses.}
  \label{fig:demo-aware-ablation}
\end{figure*}

Figure~\ref{fig:demo-aware-ablation} shows the resulting
contrast.  With the demonstration in hand, the Coach can
ground each patch on the author's resolution and produce
rules that downstream Editor runs realise more reliably, as
the marked details show.  As an empirical observation across
our runs, the two variants also tend to commit qualitatively
different kinds of rules.  Without a direct aesthetic signal,
the demonstration-free Coach has no easy way to tell which parts of
the output read well and which do not, so its proposed rules
lean toward process-level patterns such as verification
loops, restructuring checks, and panel-level reorganisation.
With the demonstration available, the Coach can compare the
Editor's output to the author's resolution at the pixel
level, and the rules it proposes become more sensitive to
rendering specifics such as math typesetting, colour, and
stroke choices.  In effect, the demonstration lowers the
evolution difficulty by letting the Coach read concrete
improvements off the author's solution rather than infer them
from trajectories alone.

\subsection{Additional Qualitative Examples}
\label{sec:appendix-more-quali}

\begin{figure*}[!htbp]
  \centering
  \includegraphics[width=\linewidth]{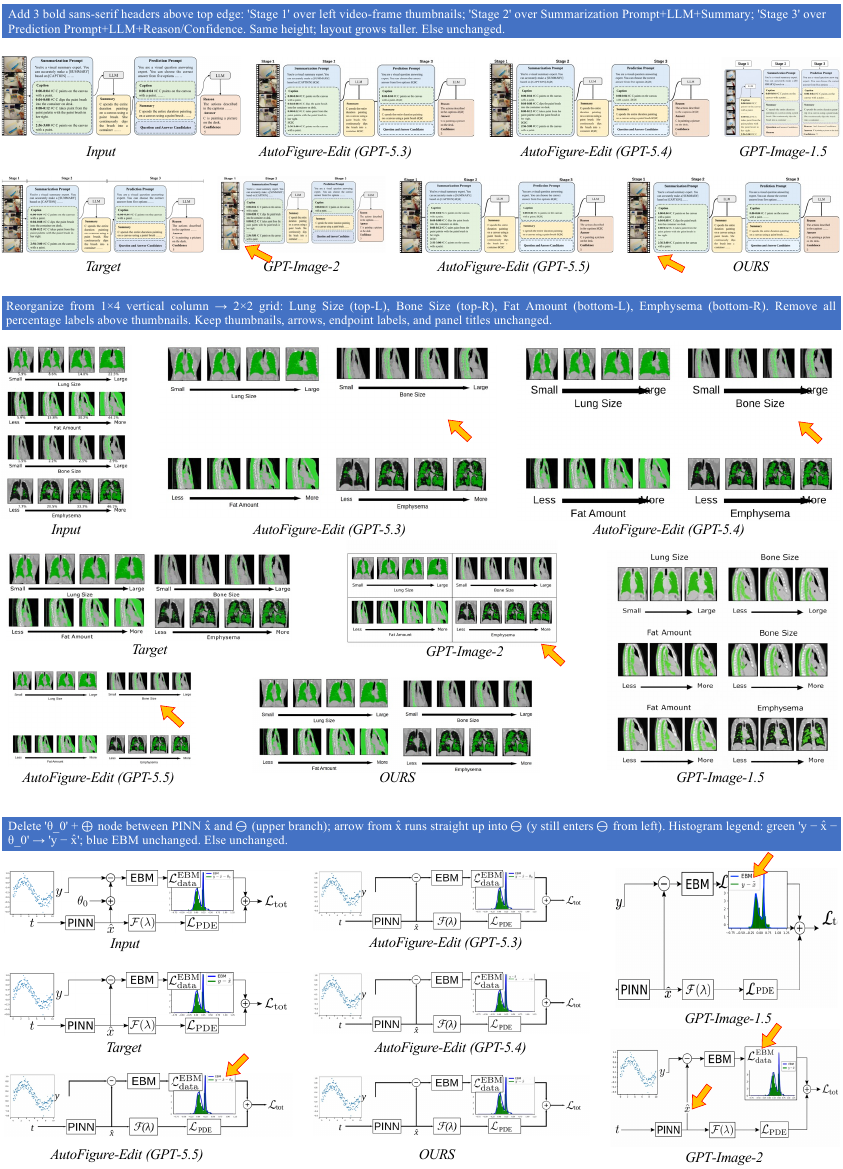}
  \caption{Additional qualitative examples.  Orange arrows
    mark notable discrepancies between methods.}
  \label{fig:more-quali-0}
\end{figure*}

\begin{figure*}[!htbp]
  \centering
  \includegraphics[width=\linewidth]{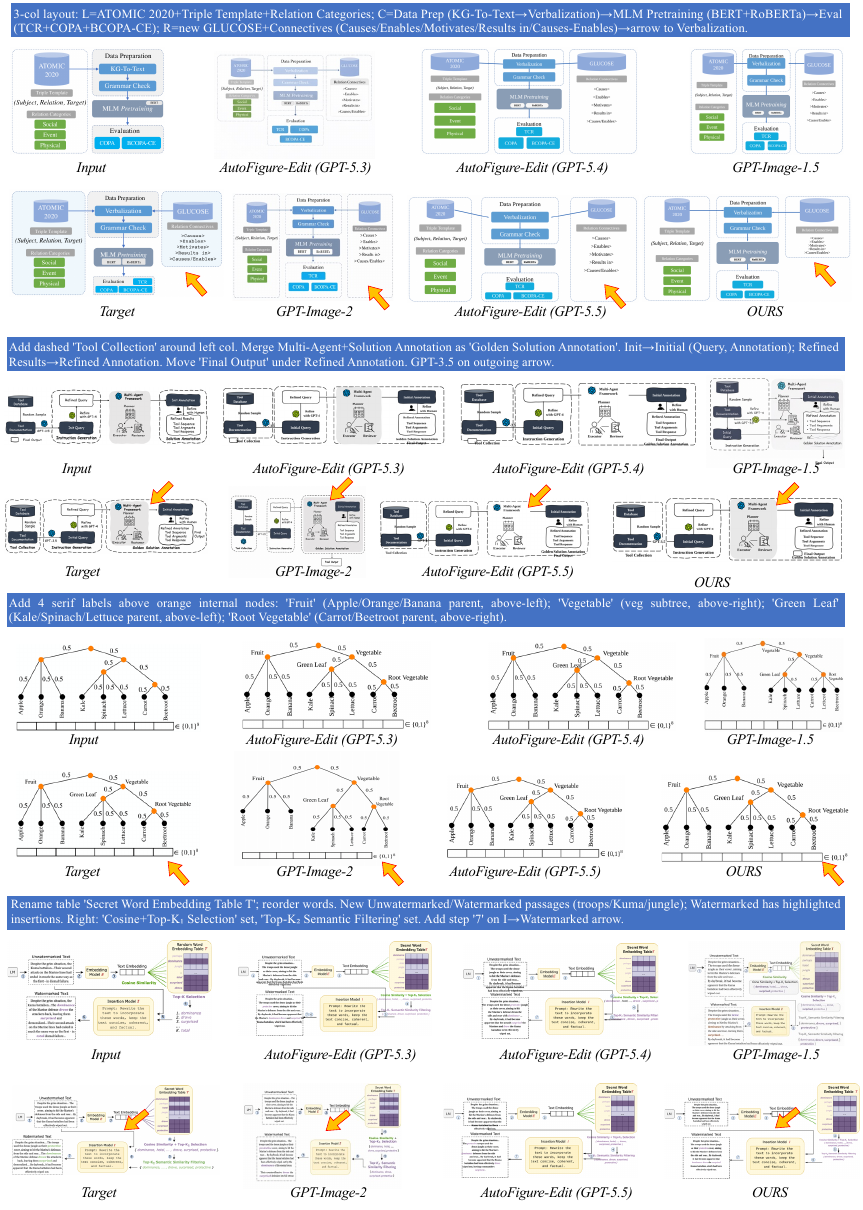}
  \caption{Additional qualitative examples.
    Orange arrows mark notable discrepancies between methods.}
  \label{fig:more-quali-1}
\end{figure*}

\begin{figure*}[!htbp]
  \centering
  \includegraphics[width=\linewidth]{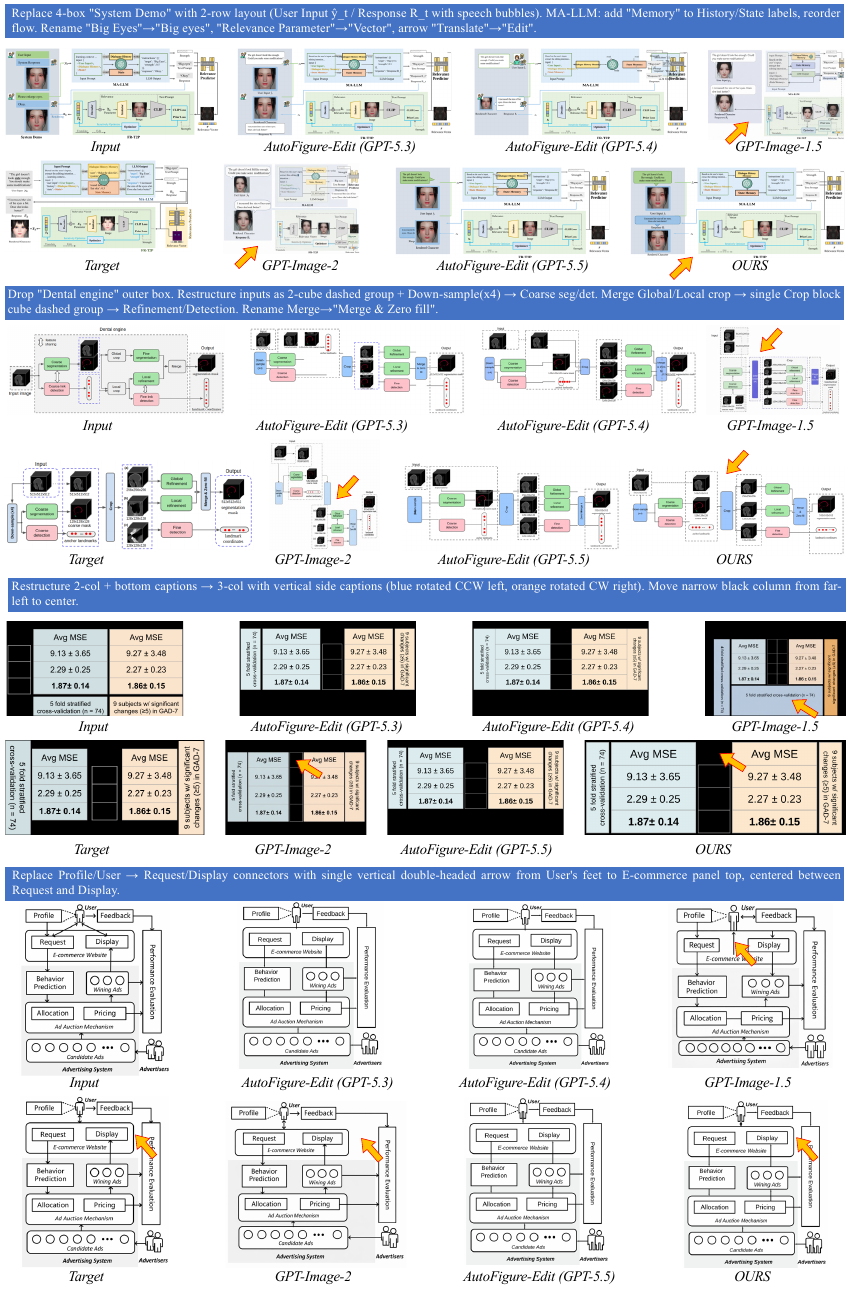}
  \caption{Additional qualitative examples.
    Orange arrows mark notable discrepancies between methods.}
  \label{fig:more-quali-2}
\end{figure*}

\begin{figure*}[!htbp]
  \centering
  \includegraphics[width=\linewidth]{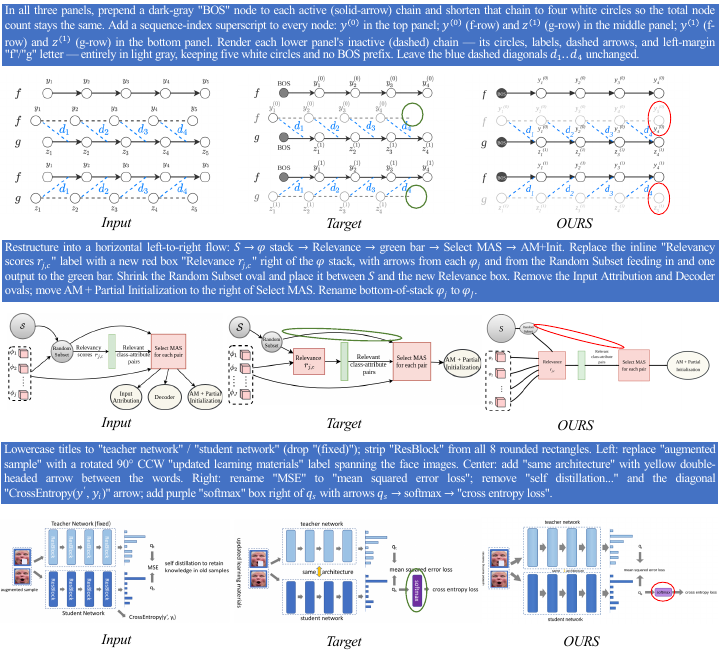}
  \vspace{-18pt}
  \caption{Three editing instances where our model captures
    the explicit instruction but misses an implicit edit the
    author also makes.  First, nodes are renumbered as
    requested, but stale nodes that the renumbering leaves
    behind are not removed.  Second, the explicit edit lands
    but an already-present connection between elements is
    dropped.  Third, a new frame is added but is not rotated
    to match the surrounding visual style.}
  \vspace{-4pt}
  \label{fig:case-study}
\end{figure*}

Figures~\ref{fig:more-quali-0}, \ref{fig:more-quali-1},
and~\ref{fig:more-quali-2} extend the qualitative comparison
in \S\ref{sec:exp-qualitative} with more editing instances
spanning different figure types and edit complexities.
Across them the same trade-off recurs.  \textsc{GPT-Image-2}
keeps individual elements visually polished but tends to lose
fidelity on the embedded text and symbols, so labels,
superscripts, and tabular numbers drift from the source even
when the surrounding layout looks plausible.  Because it
regenerates the whole figure as a raster, it also imposes its
own aesthetic prior on parts the instruction does not target,
restyling colours, fonts, and line weights that the author
left untouched.  The strongest
\textsc{AutoFigure-Edit} variant, \textsc{GPT-5.5}, preserves
the text and symbols correctly because it edits the SVG
source, but the visual register is noticeably weaker: new
shapes are drawn as crude primitives, panel positions sit
off-grid, and the added elements do not blend with the source
figure's style.  \textsc{Ours} keeps the same SVG structure
as \textsc{AutoFigure-Edit} while matching the surrounding
visual register closely enough that the output reads as a
continuation of the original figure rather than a sketch on
top of it.

\subsection{Case Study of Imperfect Outputs}
\label{sec:appendix-case-study}

Figure~\ref{fig:case-study} highlights three representative
instances where our model captures the explicit edit but
misses an implicit one that the author also makes.  These
misses point to limitations the current framework does not
yet address, such as reasoning over the figure's underlying
argument and the kind of aesthetic taste a human reader
brings to revision.

\end{document}